\documentclass[letterpaper, 10 pt, conference]{ieeeconf}  % Comment this line out if you need a4paper

\IEEEoverridecommandlockouts                              % This command is only needed if 
\usepackage{amsmath} % assumes amsmath package installed
\usepackage{amssymb}  % assumes amsmath package installed
\usepackage{svg}
\usepackage{caption}
\usepackage{multirow}
\usepackage{graphics}
\usepackage{mathtools}
\usepackage{makecell}
\usepackage[linesnumbered,ruled,vlined]{algorithm2e}

\title{\LARGE \bf
Representing 3D sparse map points and lines for camera relocalization
}

\author{Bach-Thuan Bui$^{1}$, Huy-Hoang Bui$^{1}$, Dinh-Tuan Tran$^{2}$, and Joo-Ho Lee$^{2}$% <-this % stops a space
\thanks{$^{1}$Graduate School of Information Science and Engineering, Ritsumeikan University, Japan.
        }%
\thanks{$^{2}$College of Information Science and Engineering, Ritsumeikan University, Japan.}%
}

\begin{document}

\maketitle
\thispagestyle{empty}
\pagestyle{empty}

%%%%%%%%%%%%%%%%%%%%%%%%%%%%%%%%%%%%%%%%%%%%%%%%%%%%%%%%%%%%%%%%%%%%%%%%%%%%%%%%
\begin{abstract}
Recent advancements in visual localization and mapping have demonstrated considerable success in integrating point and line features. However, expanding the localization framework to include additional mapping components frequently results in increased demand for memory and computational resources dedicated to matching tasks. In this study, we show how a lightweight neural network can learn to represent both 3D point and line features, and exhibit leading pose accuracy by harnessing the power of multiple learned mappings. Specifically, we utilize a single transformer block to encode line features, effectively transforming them into distinctive point-like descriptors. Subsequently, we treat these point and line descriptor sets as distinct yet interconnected feature sets. Through the integration of self- and cross-attention within several graph layers, our method effectively refines each feature before regressing 3D maps using two simple MLPs. In comprehensive experiments, our indoor localization findings surpass those of Hloc and Limap across both point-based and line-assisted configurations. Moreover, in outdoor scenarios, our method secures a significant lead, marking the most considerable enhancement over state-of-the-art learning-based methodologies. The source code and demo videos of this work are publicly available at: https://thpjp.github.io/pl2map/. 

\end{abstract}

%%%%%%%%%%%%%%%%%%%%%%%%%%%%%%%%%%%%%%%%%%%%%%%%%%%%%%%%%%%%%%%%%%%%%%%%%%%%%%%%
\section{INTRODUCTION}
\label{section1}
Owing to the cost-effectiveness and abundant texture resources of visual features, their application in localization and mapping has gained considerable traction in the realms of robotics and computer vision. In contrast to mere point-based methods, incorporating line-assisted features offers a deeper understanding of scene layouts and geometric cues, paving the way for more versatile and efficient applications.

Recent studies have shown that simultaneous localization and mapping (SLAM) or structure from motion (SfM) performance can be enhanced by integrating both points and lines \cite{xu2023airvo, liu20233d, zhang2023pl, gomez2019pl, shu2023structure}. However, localization based on maps pre-built using SLAM or SfM-based methods often requires huge computational resources for feature matching (FM) between local images and global maps \cite{sarlin2019coarse, liu20233d}. This process requires the storage of pre-built 3D maps \cite{sarlin2019coarse, sarlin2020superglue}, including detailed descriptor components, which proves to be prohibitively costly for real-time applications, particularly in the context of applications with lightweight robotics platforms. 

\begin{figure}
    \centering
    \includegraphics[width=0.45\textwidth]{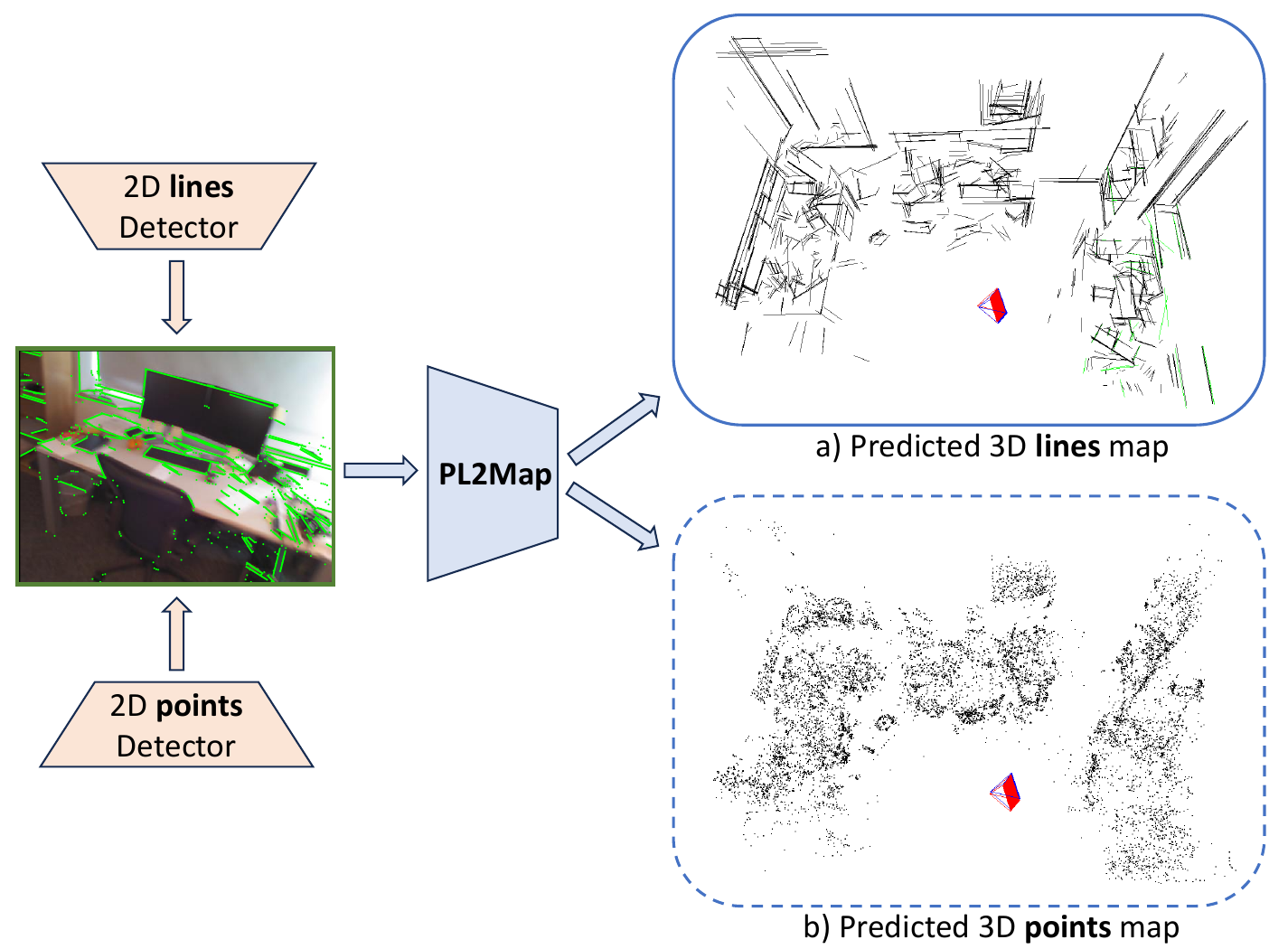}
    \caption{\textbf{Representing 3D point-line maps by PL2Map}. We show an example of the results of the proposed learning method for representing 3D point-line features. The red camera poses in both predicted lines (a) and points (b) map are the ground truth poses of the input image on the left, and the blue ones are the estimated camera poses using predicted lines or points map.}
    \label{fig:enter-label}
\end{figure}

Therefore, in this study, we introduce PL2Map, a novel neural network tailored for the efficient representation of complex point and line maps. This methodology naturally provides 2D-3D correspondences, simplifying the relocalization task by foregoing the need for expensive feature matching and descriptor management. 

Our method aims to map sparse descriptors directly to 3D coordinates using a neural network, which can encounter several challenges. The primary issue is the variability in the number of points and lines, which fluctuates with changes in the viewpoint. Furthermore, the sequence in which points and lines are arranged is inconsistent, and there is always variability in line lengths \cite{yoon2021line, pautrat2021sold2}. This introduces additional complexity in accurately mapping 2D sparse points and lines to 3D space compared with more uniform approaches \cite{brachmann2021visual, brachmann2023accelerated, yang2019sanet, zhou2020kfnet}

To overcome the aforementioned challenges, we first drew inspiration from the principles of feature matchers \cite{sarlin2020superglue, lindenberger2023lightglue}, treating points and lines as two distinct yet interrelated sets of unordered descriptors. To account for variations in line lengths and ensure their unique features, we adopted a strategy inspired by \cite{yoon2021line}, conceptualizing lines as sequences of words, with each word representing an intra-point descriptor. We then used a transformer encoder model \cite{vaswani2017attention} to encode each line sentence as a unique point-like descriptor, thereby streamlining the line descriptor extraction process within the PL2Map's preprocess, allowing for a shared extractor for both points and lines. Subsequently, we utilized self and cross-attention mechanisms within several graph layers to facilitate the exchange and refinement of feature descriptors. Following this attention-driven update, the points and line features were split into two separate Multilayer Perceptrons (MLPs) to regress their respective 3D coordinates. The contributions of this study are as follows.
\begin{itemize}
    \item To the best of our knowledge, our method of direct learning in mapping 2D-3D correspondences for point and line features represents this first attempt at camera relocalization. 
    \item We propose a complete learning pipeline including network architecture, and robust loss functions for learning to represent both points and lines from pre-built SfM models. Through the proposed end-to-end training pipeline, the maps of points and lines can be further refined, leading to improvements in subsequent camera relocalization. 
    \item We set a new record on two localization benchmarks of the 7scenes \cite{shotton2013scene} and Cambridge Landmarks \cite{kendall2015posenet}, in which, our PL2Map surpasses both FM-based Hloc and Limap performance on 7scenes. For outdoor Cambridge Landmarks, our pipeline marks the most significant enhancement over state-of-the-art learning-based approaches. 
\end{itemize}

\section{RELATED WORK}
\subsection{Image Retrieval and Pose Regression}
Image retrieval-based methods relocalize by simply comparing a query image with posed images in the database \cite{arandjelovic2016netvlad} and approximating the pose of the query with the highest-ranking retrieved images \cite{sattler2019understanding}. Pose regression employs a neural network to predict the absolute camera pose from a query image or the relative pose between a query and retrieved images \cite{kendall2015posenet, bach2022featloc, zhou2020kfnet, sattler2019understanding, zhou2020learn}. However, both image retrieval and pose regression-based methods exhibit lower accuracy than the image structure-based methods discussed subsequently. 

\subsection{Feature Matching} 
Feature-matching-based (FM) approaches are typically divided into direct \cite{svarm2016city, zeisl2015camera} and indirect methods \cite{sarlin2019coarse, liu20233d, gao2022pose}. Direct methods employ a strategy of matching 2D-3D correspondences directly from query image features to 3D points in SfM models. Although direct approaches can yield accurate camera poses, they are limited in scalability to larger scenes because of memory consumption and ambiguities \cite{sarlin2019coarse}. Conversely, indirect approaches begin with an image retrieval step against images in the database \cite{sarlin2019coarse, liu20233d} and then match the features from the query image to those in the retrieved images. This creates a very robust pipeline under challenging conditions \cite{sarlin2020superglue, lindenberger2023lightglue}.

Traditionally, FM-based methods rely solely on point features for relocalization. However, recent advancements suggest that incorporating line features can significantly enhance this pipeline \cite{gao2022pose, liu20233d, xu2023airvo}. However, the introduction of additional features increases the complexity and cost of FM-based pipelines owing to the additional matching steps required for line features \cite{pautrat2023gluestick, yoon2021line, pautrat2021sold2}, and additional memory is required for storing line descriptors \cite{liu20233d}.  

\subsection{Learning Surrogate Maps}
Another prominent category within relocalization methodologies is scene coordinate regression (SCR) \cite{shotton2013scene, brachmann2021visual, zhou2020kfnet, brachmann2023accelerated}. These techniques infer 3D coordinates within the scene space from dense 2D pixel positions in the images, effectively embedding map representations within the neural network's weights. This approach offers notable benefits, including minimal storage requirements \cite{brachmann2023accelerated, bui2022fast} and enhanced privacy due to the implicit nature of the map representation \cite{speciale2019privacy}. Nevertheless, recent learning-based approaches \cite{bui2023d2s, do2022learning, nguyen2024focustune} have demonstrated that focusing on key landmarks can improve relocalization accuracy and robustness to environmental changes \cite{bui2023d2s, do2022learning}. Specifically, the sparse map learning approach detailed in \cite{bui2023d2s} has shown superior performance compared with dense SCR \cite{zhou2020kfnet, brachmann2021visual}, particularly in scenarios characterized by significant domain shifts or limited training data availability.

Our method aligns closely with this innovative trend as we endeavor to create neural-based surrogate maps that incorporate both point and line features. This, results in significant enhancement of the camera relocalization task.

\section{PROPOSED METHOD}

\begin{figure*}
    \centering
    \includegraphics[width=\textwidth]{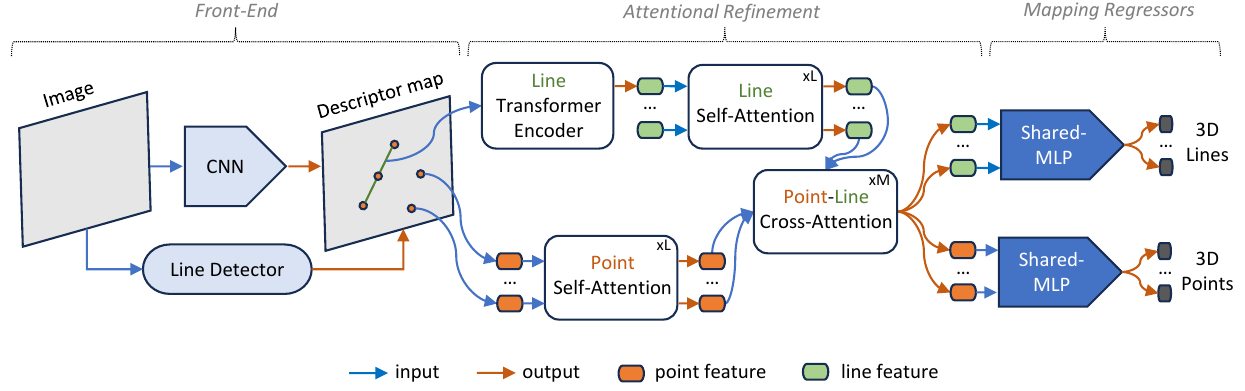}
    \caption{\textbf{PL2Map pipeline}. We illustrate the architecture of PL2Map, which consists of three main components: \textit{Front-End}, \textit{Attentional Refinement}, and \textit{Mapping Regressors}.}
    \label{whole_pipeline_network}
\end{figure*}

\subsection{Problem Statements}
Recent advancements in SfM and visual SLAM have been explored with many successful mapping elements, such as points, lines, edges, planes, and objects \cite{xu2023airvo, shu2023structure, shu2021visual, liu20233d, zhang2023pl, wu2023object, liao2022so}. With the demand for additional mapping elements, there is a clear need for a more efficient mapping representation strategy that extends beyond the basic storage of descriptor vectors \cite{sarlin2020superglue, sarlin2019coarse, liu20233d}. To address this issue, specifically for point and line maps, we introduce a neural-based surrogate model capable of representing both 3D points and lines through their descriptors. This simplifies the matching process for multiple mapping elements. 

Assume that we have a set of 2D keypoints $\{\mathbf{p}_{i}\}^{N}$ and a set of 2D line segments $\{\mathbf{l}_{i}\}^{M}$ extracted from image $\mathbf{I}^{r}$, each associated with visual descriptors $\{\mathbf{d}_{i}^{p}\}^{N}$  and $\{\mathbf{d}_{i}^{l}\}^{M}$ respectively. Here, $r$ denotes the image sourced from the reference database used to construct the 3D points and line map. We aim to develop a learning function $\mathcal{F}(.)$ that inputs the two sets of \textit{ visual descriptors} $\{\mathbf{d}_{i}^{p}\}^{N}$ and $\{\mathbf{d}_{i}^{l}\}^{M}$, and outputs the corresponding sets of 3D points $\{\mathbf{P}_{i}\in \mathbb{R}^{3}\}^{N}$ and lines $\{\mathbf{L}_{i} \in \mathbb{R}^{6}\}^{M}$ sets in the world coordinates system. The ultimate goal is to estimate a six degrees of freedom (6 DOF) camera pose $\mathbf{T} \in \mathbb{R}^{4\times4}$ for any new query image $\mathbf{I}$ from the same environment.

\subsection{PL2Map}
This section presents in detail the PL2Map model designed to learn the representation of the sparse 2D-3D correspondences for both points and lines. Both points and lines possess interchangeable features, such as line endpoints and adjacent points, which can be integrated to enhance the development and accuracy of the final 3D map \cite{pautrat2023gluestick}. Fig. \ref{whole_pipeline_network} shows the proposed architecture including three sub-blocks.
\begin{itemize}
    \item \textit{Front-End}: We use the available 2D detectors to extract both the point and line positions and their descriptors from the input image.  
    \item \textit{Attentional Refinement}: Points and line features/descriptors are boosted by a subsequent attentional refinement module, targeting the awareness of the surrounding point and line features \cite{bui2023d2s}. 
    \item \textit{Mapping Regressors}: The last module consists of two MLPs that are used to regress the 3D map lines and points. 
\end{itemize}

\subsubsection{Front-End}
The proposed system inputs the available 2D point and line descriptor sets of $\{\mathbf{d}^{p}\}^{N}$ and $\{\mathbf{d}^{l}\}^{M}$, extracted from the images. To gather these inputs, we rely on off-the-shelf 2D point and line detectors, either hand-crafted strategies or learning-based approaches, such as SIFT \cite{lowe2004distinctive}, SuperPoint \cite{detone2018superpoint}, LSD \cite{von2008lsd}, and DeepLSD \cite{pautrat2023deeplsd}. 

For the descriptor features of the points, we utilized the direct results produced by the extractors. For lines, rather than employing separate line descriptors, we opted to use point descriptors to represent the lines. This approach is more convenient and cost-effective for subsequent inference processes. To achieve this, we uniformly sampled $T$ point descriptors to represent a 2D line, which subsequently served as the input for the attentional refinement module. This sampling process is shown on the left-hand side of Fig. \ref{line_transformer}. 
\subsubsection{Attentional Refinement}
\textit{Attentional refinement} is a key component of our method and is comprised of three submodules: Line Transformer Encoders, Self-Attention, and Cross-Attention. Each submodule is specifically designed to augment the features of lines and points by utilizing descriptor similarity. 

\begin{figure}
    \centering
    \includegraphics[width=0.5\textwidth]{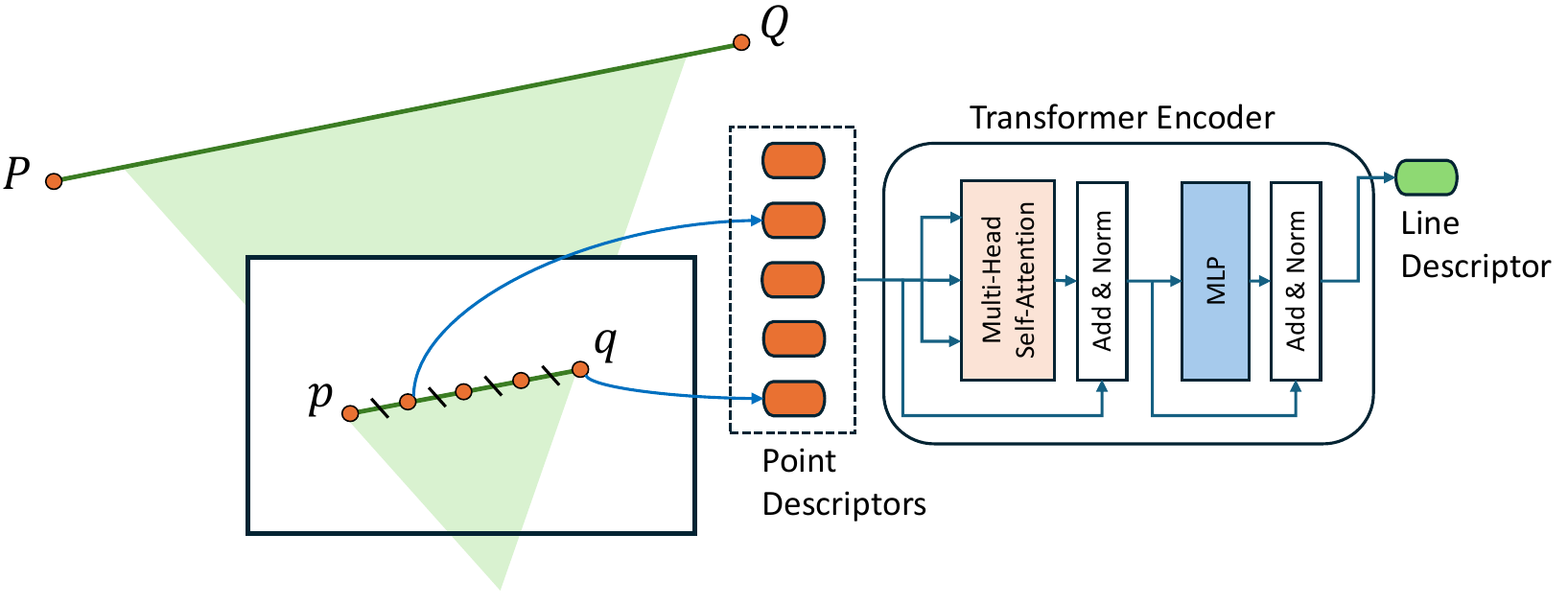}
    \caption{\textbf{Line transformer encoder}. We represent a 2D local line by uniformly sampling $T-2$ number of points inside the line segment of endpoints $p$ and $q$. A transformer-based model is then used to uniformly transform all point descriptors to a single feature with the same dimension, which can be considered as a line descriptor.}
    \label{line_transformer}
\end{figure}

\textbf{Line Transformer Encoder}. Because the point and line features extracted from the \textit{front-end} module have different dimensions, we initially employed a transformer-based encoder \cite{vaswani2017attention} to align the dimensions of the line descriptors similar to those of the points. The process for encoding line descriptions is shown in Fig. \ref{line_transformer}. In our approach, a line segment is characterized by two endpoints, $p$ and $q$. Similar to \cite{yoon2021line}, we uniformly sampled $T-2$ interval point tokens and their descriptors for a line segment. This process yields an embedded line token $\mathbf{L} \in \mathbb{R}^{T\times D}$, where $D$ represents the descriptor dimension, which is consistent with that of the point descriptors. 

Unlike the approach in \cite{yoon2021line}, we did not incorporate the position and orientation of the line in our method. Instead, the input to the transformer model consisted solely of point descriptors, as illustrated in Fig.\ref{line_transformer}. This decision was based on the observation that a line's appearance in terms of position and orientation can vary significantly with changes in the camera view, whereas its 3D position remains constant in the world coordinate system. Previous studies, such as \cite{bui2023d2s}, demonstrated that including positional factors in 3D regression modules can lead to suboptimal performance in mapping regression. We assume that the descriptors of interval points along a corresponding segment follow a similar pattern. The 3D coordinates ($PQ$) of the 2D line ($pq$), as shown in Fig. \ref{line_transformer}, are also illustrated with different reprojection lengths but can be represented by using only 2D sampled points. 

To embed the line tokens of $\mathbf{L}$ to have the same dimensions as the point descriptor, we use a single transformer model $\mathfrak{T}$, which can be written as follows: 

\begin{equation}
\begin{aligned}
\mathfrak{T}(\mathbf{L}) & = \mathbf{d}^{l},%\\
 % \mathbf{L}_{1} & = NL(MSA(\mathbf{L})+\mathbf{L}), \\
 % \mathbf{L}_{2} & = NL(MLP(\mathbf{L}_{1})+\mathbf{L}_{1}), 
\label{ori_equation}
\end{aligned}
\end{equation}

where $\mathfrak{T}:\mathbb{R}^{N \times D} \to \mathbb{R}^{1 \times D}$. The architecture of $\mathfrak{T}$ is shown in Fig. \ref{line_transformer}, which consists of two main components: a multi-head self-attention layer and an MLP layer, whereas each sub-layer is appended by a residual connection and layer normalization. 

\textbf{Self and Cross Attention}. Similar to previous studies \cite{sarlin2020superglue, bui2023d2s, pautrat2023gluestick}, we also consider the attention module as a complete graph with two types of undirected edges. The self-attention edge $\mathcal{E}_{self}$ connects all the surrounding descriptors of either points or lines in the same image, whereas the cross-attention edge $\mathcal{E}_{cross}$ connects points to lines and lines to points. 

Let $\prescript{(m)}{}{\mathbf{d}_{i}}$ be the intermediate descriptor for element $i$ in layer $m$. We initialized $\prescript{(0)}{}{\mathbf{d}_{i}} = \mathbf{d}_{i}$. Then, the residual update for all $i$ is:

\begin{equation}
\begin{aligned}
\prescript{(m+1)}{}{\mathbf{d}_{i}} = \prescript{(m)}{}{\mathbf{d}_{i}} + \phi_{m}\bigg(\bigg[\prescript{(m)}{}{\mathbf{d}_{i}}|| a_{m}\big(\prescript{(m)}{}{\mathbf{d}_{i}}, \mathcal{E}\big) \bigg]\bigg)
\label{ori_equation}
\end{aligned}
\end{equation}

where $\mathcal{E} \in \{\mathcal{E}_{self}, \mathcal{E}_{cross}\}$, $[.||.]$ denotes the concatenation, $\phi_{m}$ is modeled with an MLP, and $a_{m}\big(\prescript{(m)}{}{\mathbf{d}_{i}}; \mathcal{E}\big)$ is the Multi-Head Attention from \cite{vaswani2017attention} applied to the set of edges $\mathcal{E}$, which is calculated as follows: 

\begin{equation}
\begin{aligned}
a_{m}(\prescript{(m)}{}{\mathbf{d}_{i}}, \mathcal{E}) = \sum_{j:(i,j)\in\mathcal{E}}{\alpha_{ij}\mathbf{v}_{j}}
\label{ori_equation}
\end{aligned}
\end{equation}

where $\alpha_{ij}=$ Softmax$_{j}\big(\mathbf{q}_{i}^{T}\mathbf{k}_{j}/\sqrt{D/h}\big)$ is the attention score, query $\mathbf{q}_{i}$ key $\mathbf{k}_{j}$, and value $\mathbf{v}_{j}$ are the linear projections of descriptors $\mathbf{d}_{i}$ and $\mathbf{d}_{j}$, and $h$ is the number of heads. In self-attention $\mathbf{k}_{j}$ and $\mathbf{v}_{j}$ come from the same descriptor set with $\mathbf{q}_{i}$ (either points or lines), whereas in cross-attention, if $\mathbf{q}_{i}$ comes from the points set, $\mathbf{k}_{j}$ and $\mathbf{v}_{j}$ will be calculated using the line descriptors set, and vice versa.

\subsubsection{Mapping Regressors}
Finally, we use two different MLP networks to regress the 3D coordinates of the points and lines. The models input the \textit{fine descriptors} resulting from the attentional module as follows:

\begin{equation}
\begin{aligned}
\hat{\mathbf{P}}_{i}= \phi^{p}\big( \prescript{(m)}{}{\mathbf{d}^{p}} \big)
\label{ori_equation}
\end{aligned}
\end{equation}
\begin{equation}
\begin{aligned}
\hat{\mathbf{L}}_{i} = \phi^{l}\big( \prescript{(m)}{}{\mathbf{d}^{l}} \big)
\label{ori_equation}
\end{aligned}
\end{equation}

where $\phi^{p}:\mathbb{R}^{D} \to \mathbb{R}^{4}$ and $\phi^{l}:\mathbb{R}^{D} \to \mathbb{R}^{7}$ are shared by all descriptors in the same set (points or lines). Because the number of triangulated 3D points and lines is different from the 2D points and lines detected from every image, we extend one more dimension for reliability prediction of $\hat{\mathbf{P}}_{i}$ and $\hat{\mathbf{L}}_{i}$, which is similarly defined in \cite{bui2023d2s}, as follows: 

\begin{equation}
\begin{aligned}
\hat{r} = \frac{1}{1+|\beta z|} \in (0,1],
\end{aligned}
\label{reliable_equation}
\end{equation}

where $z$ is the last value of $\hat{\mathbf{P}_{i}} = (\hat{\mathbf{P}}_{i}^{'}, z)^{T}$ or $\hat{\mathbf{L}}_{i} = (\hat{\mathbf{L}}_{i}^{'}, z)^{T}$, $\beta$ is a scale factor chosen to make the expected reliability $r$ easy to reach a small value when input descriptors have no 3D coordinates.

\begin{figure}
    \centering
    \includegraphics[width=0.25\textwidth]{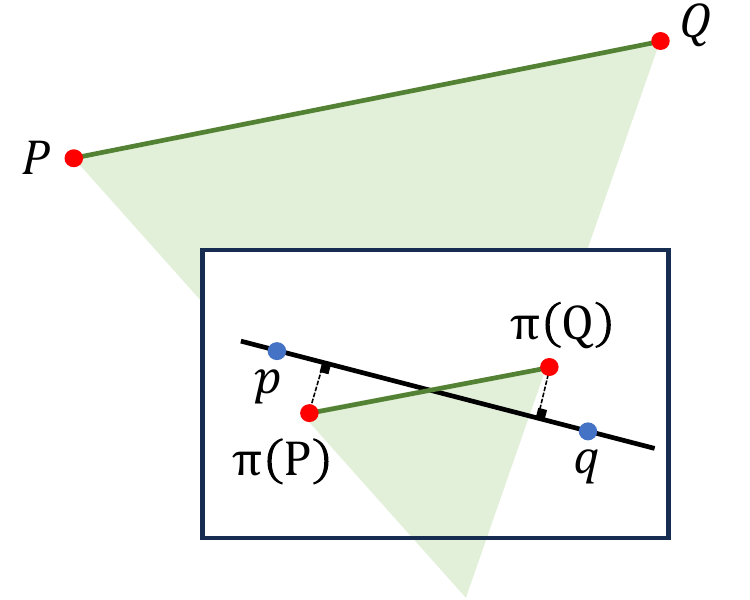}
    \caption{\textbf{Line reprojection loss}. Given two 2D endpoints $p$ and $q$, and their predictions of 3D endpoints $P$ and $Q$, we minimize the reprojection distance of $\pi(P)$ and $\pi(Q)$ to the 2D segment $pq$ on the image plane. This allows the length of $PQ$ in 3D space independent with 2D segment $pq$ length, which can also solve the occlusion problem in the camera view.}
    \label{fig:line_reprojection}
\end{figure}

\subsection{Loss Function}
\begin{figure}
    \centering
    \includegraphics[width=0.5\textwidth]{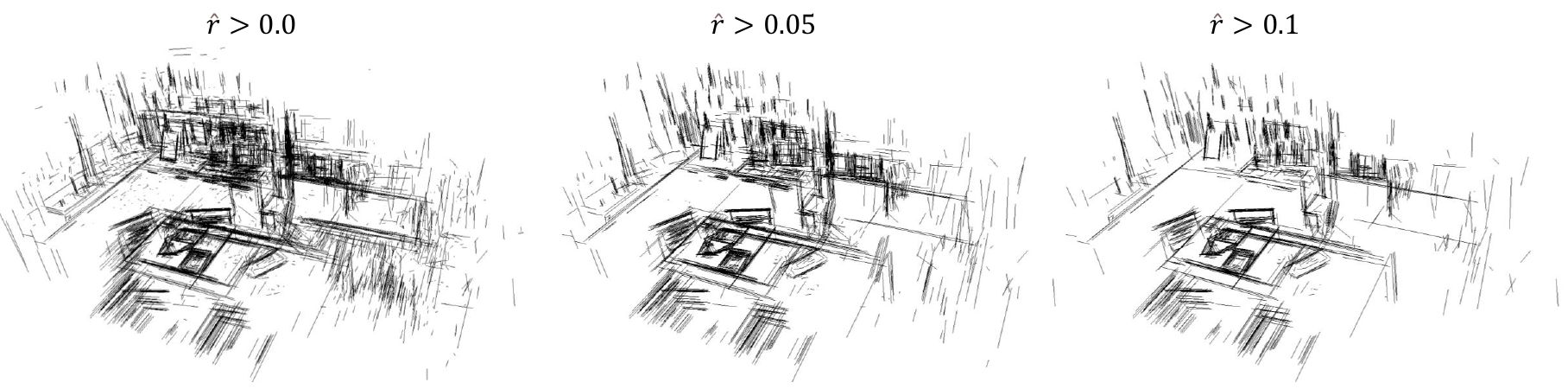}
    \caption{\textbf{Reliable Line-Map Prediction Results.} We show predicted line-map filtering with a different threshold $\hat{r}$ in RedKitchen scene from 7scenes \cite{shotton2013scene}}
    \label{reliable_filter_linemap}
\end{figure}

The predicted $\hat{\mathbf{P}_{i}}$ and $\hat{\mathbf{L}}_{i}$ are optimized using their pseudo ground truths $\mathbf{P}_{i}$ and $\mathbf{L}_{i}$ from 3D models for each image as follows:

\begin{equation}
    \mathcal{L}_{m} = \sum_{i=1}^{N} r_{i}^{p}\lVert \mathbf{P}_{i}-\hat{\mathbf{P}}_{i}^{'} \lVert_{\gamma} + \sum_{i=1}^{M} r_{i}^{l}\lVert \mathbf{L}_{i}-\hat{\mathbf{L}}_{i}^{'} \lVert_{\gamma},
\label{loss_m}
\end{equation}

where $r\in\{0,1\}$, $\gamma$ is a robust norm. Because the loss Eq. \ref{loss_m} only focuses on regressing valid points and lines, we optimize the reliability prediction for non-robust descriptors as follows: 

\begin{equation}
    \mathcal{L}_{r} = \sum_{i=1}^{N}  \lVert r_{i}^{p}- \hat{r}_{i}^{p} \lVert_{\gamma} + \sum_{i=1}^{M}  \lVert r_{i}^{l}- \hat{r}_{i}^{l} \lVert_{\gamma}.
\end{equation}

where $\hat{r}_{i}$ is calculated using Eq. \ref{reliable_equation} for both points and lines. In Fig. \ref{reliable_filter_linemap}, we show an example of reliability prediction results. Furthermore, we optimize the model using the available camera poses by reprojecting the predicted 3D points and lines onto the image plane: 

\begin{equation}
\begin{aligned}
    \mathcal{L}_{\pi} = & \sum_{i=1}^{N} v_{i}^{p} r_{i}^{p}\big\lVert \pi(\mathbf{T},\mathbf{P}_{i})-\mathbf{u}_{i}^{p}\big\lVert_{\gamma} \\ & + \sum_{i=1}^{M} v_{i}^{l} r_{i}^{l} \psi\big(\pi(\mathbf{T},\mathbf{L}_{i}), \mathbf{u}_{i}^{l}\big),
\label{origin_reproject_loss}
\end{aligned}
\end{equation} 

where $\mathbf{T}$ is the ground truth pose, $\pi(.)$ is the reprojection function, $\mathbf{u}_{i}^{p} \in \mathbb{R}^{2}$ and $\mathbf{u}_{i}^{l} \in \mathbb{R}^{4}$ are the 2D positions of the point and line endpoints on the image, $\psi(.)$ is a function that calculates the distance between reprojected 3D lines and its ground truth coordinates $\mathbf{u}_{i}^{l}$, as illustrated in Fig. \ref{fig:line_reprojection}, and $v_{i} \in \{0,1\}$, $v_{i}=1$ when a prediction is between 10 cm and 1000 m in front of the camera, and has projection error lower than 1000 px, and otherwise $v_{i}=0$.  

However, the reprojection loss, as illustrated in Eq. \ref{origin_reproject_loss}, is still highly non-convex and difficult to train at the beginning stage. Thus, we adapt the robust projection error defined in \cite{brachmann2023accelerated} as follows: 

\begin{equation}
\begin{aligned}
    \mathcal{L}_{\pi}^{robust} = \tau(t) tanh\bigg(\frac{\mathcal{L}_{\pi}}{\tau(t)}\bigg),
\label{reproject_loss}
\end{aligned}
\end{equation} 

where $\tau(.)$ is the threshold used to dynamically rescale $tanh(.)$, and varies throughout the training:

\begin{equation}
\begin{aligned}
    \tau(t) = \omega(t) \tau_{max} + \tau_{min},  \text{ with } \omega(t) = \sqrt{1-t^{2}}, 
\label{reproject_loss}
\end{aligned}
\end{equation} 

where $t\in(0,1)$ denotes the relative training progress. This forces the threshold $\tau$ to have a circular schedule that remains close $\tau_{max}$ at the beginning and reaches $\tau_{min}$ at the end of the training. 

Finally, we integrate all loss functions to optimize the surrogate model as follows: 

\begin{equation}
    \mathcal{L}= \delta_{m}\mathcal{L}_{m} + \delta_{r}\mathcal{L}_{r} + \delta_{\pi}\mathcal{L}_{\pi}^{robust},
    \label{final_loss}
\end{equation}

where $\delta$ is the hyperparameter coefficient used to balance three loss functions.

\begin{table*}
\centering
\caption{\textbf{Localization results on 7scenes \cite{shotton2013scene}}. We report the median translation and rotation errors in cm and degrees and pose accuracy (\%) at 5 cm / 5 deg. threshold of different relocalization methods using points and lines on the \textit{7scenes} dataset. The methods marked with $^{\star}$ are FM-based or database-based methods. The results in \textcolor{red}{red} are the best and \textcolor{blue}{blue} indicates the second best.}
\label{results_7scenes_table}

\begin{tabular}{c|c|c|c|c|c} 
\hline\hline
           & $^\star$Hloc $^\text{point}$ \cite{sarlin2019coarse,sarlin2020superglue}                         & $^\star$PtLine $^\text{point \& line}$ \cite{gao2022pose} & $^\star$Limap $^\text{point \& line} \cite{liu20233d} $ & Pl2Map $^\text{point}$ \textbf{(ours) } & Pl2Map $^\text{point \& line}$ \textbf{(ours)}  \\ 
\hline
Chess      & 2.4 / 0.84 / 93.0                                    & 2.4 / 0.85 / 92.7                      & 2.5 / 0.85 / 92.3                      & \textcolor{blue}{2.0 / 0.65 / 95.5}     & \textcolor{red}{1.9 / 0.63 / 96.0}              \\
Fire       & 2.3 / 0.89 / 88.9                                    & 2.3 / 0.91 / 87.9                      & 2.1 / 0.84 / \textcolor{red}{95.5}     & \textcolor{blue}{2.0 / 0.81 / }93.3     & \textcolor{red}{1.9 / 0.80 /} \textcolor{blue}{94.0}              \\
Heads      & \textcolor{red}{1.1} / 0.75 / \textcolor{blue}{95.9} & \textcolor{blue}{1.2} / 0.81 / 95.2    & \textcolor{red}{1.1 }/ 0.76 / 95.9     & \textcolor{blue}{1.2 / 0.74 / 97.8}     & \textcolor{red}{1.1 / 0.71 / 98.2}              \\
Office     & 3.1 / 0.91 / 77.0                                    & 3.2 / 0.96 / 74.5                      & 3.0 / 0.89 / 78.4                      & \textcolor{blue}{2.8 / 0.78 / 82.3}     & \textcolor{red}{2.7 / 0.74 / 84.3}              \\
Pumpkin    & 5.0 / 1.32 / 50.4                                    & 5.1 / 1.35 / 49.0                      & 4.7 / 1.23 / 52.9                      & \textcolor{blue}{3.5 / 0.96 / 63.1}     & \textcolor{red}{3.4 / 0.93 / 64.1}              \\
RedKitchen & 4.2 / 1.39 / 58.9                                    & 4.3 / 1.42 / 58.0                      & 4.1 / 1.39 / 60.2                      & \textcolor{blue}{3.8 / 1.13 / 66.7}     & \textcolor{red}{3.7 / 1.10 / 68.9}              \\
Stairs     & 5.2 / 1.46 / 46.8                                    & \textcolor{blue}{4.8 / 1.33 / 51.9}    & \textcolor{red}{3.7 / 1.02 / 71.1}     & 8.5 / 2.4 / 27.8                        & 7.6 / 2.0 / 33.3                                \\
\hline\hline
\end{tabular}

\end{table*}

\section{EXPERIMENTS}
\textit{Network Setting}. We implemented our approach in Pytorch \cite{paszke2017automatic}, with the following settings for the network architecture. The graph attention consists of a single line-transformer model for all lines that receive $T=12$ point descriptors including two endpoints, and five graph attention layers of (\textit{self, cross, self, cross, self}). We used the same number of attention heads for both the transformer-encoder and self-cross-attention modules. For the final mapping layers, two different MLPs ($D$, 512, 1024, 512, 4) and ($D$, 512, 1024, 512, 7) are used to regress the 3D points and lines. 

\textit{Hyperparameters Choices}. We chose $\beta=100$ and initialized the soft threshold $\tau_{max}= \text{50px}$ for training with the indoor dataset, $\tau_{max}= \text{100px}$ for the outdoor dataset, and $\tau_{min}= \text{1px}$. The hyperparameters $\delta$ in Eq. \ref{final_loss} were selected as one for all three sub-losses. We optimized our method using Adam \cite{kingma2014adam} with a start learning rate of $3\cdot10^{-4}$ and shrank it seven times with a decay parameter of 0.5, for the indoor dataset. For outdoors, we use a smaller learning rate of $5\cdot10^{-5}$ and shrank it ten times. We then trained all the environments with 2.5M iterations. 

We applied data augmentation to all the experiments. Specifically, we applied random adjustments to the brightness $\pm15\%$ and contrast $\pm10\%$ of the input image. We further randomly rotated the image $\pm30^{\circ}$ and re-scaled the images within 66 and 150 percent. For each applied augmentation, we adjusted the camera poses and focal lengths according to the changes in the images. 
All experiments were performed using an Nvidia GeForce GTX 1080ti GPU and Intel Core i9-9900 CPU.

\textit{Camera Pose Estimation}. Given the predicted points and line maps, we estimated the camera poses using two different settings, one using points only, and another using both points and lines for visual localization. For points, we directly used RANSAC PnP implemented by Hloc \cite{sarlin2019coarse}. For points and lines-assisted, we used the same mechanism as in Limap \cite{liu20233d}. 

\begin{figure*}
    \centering
    \includegraphics[width=\textwidth]{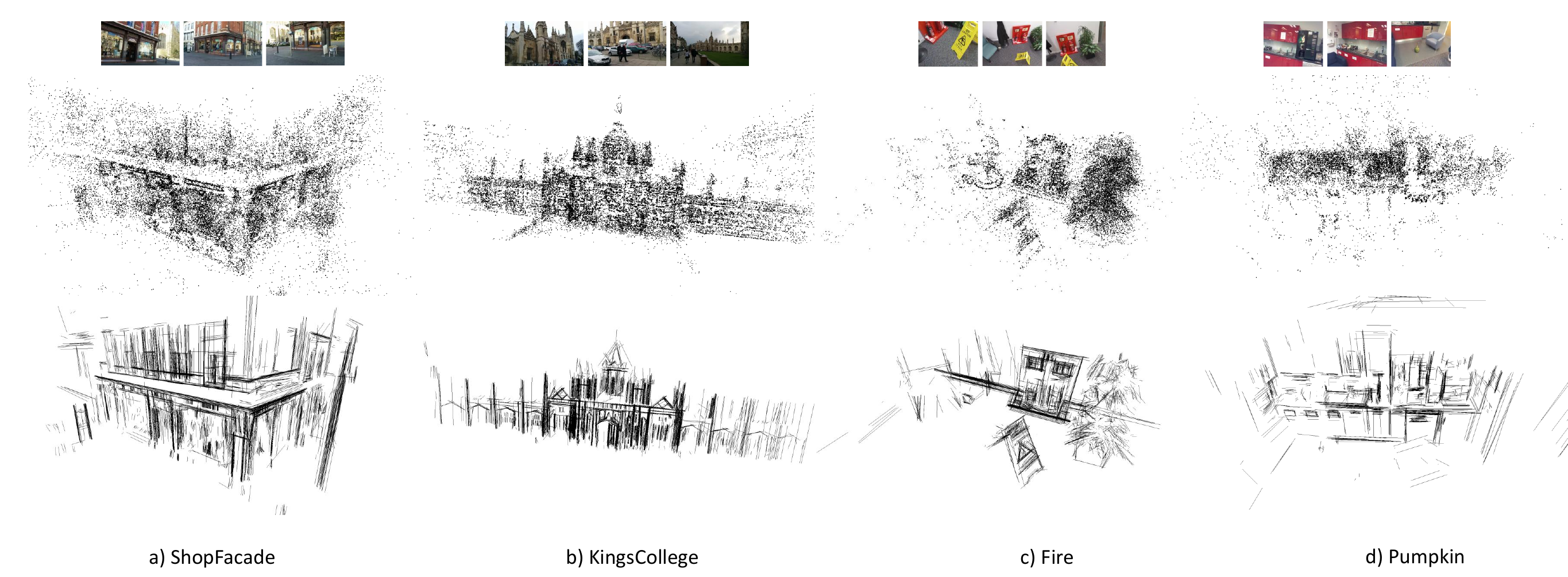}
    \caption{Qualitative results on both outdoor and indoor scenes (a, b from the Cambridge dataset \cite{kendall2015posenet} and c, d from the 7scenes dataset \cite{shotton2013scene}). Three different views are shown on the top. The second and third rows are our prediction results of 3D point and line maps respectively, using a random number of 20 test images.}
    % $\hat{r}^{l}=0.2$ for outdoor and 0.05 for indoor, $\hat{r}^{p}=0.6$ for all
    \label{predicted_3d_map}
\end{figure*}

\begin{table}
\centering
\caption{\textbf{Localization results on 7scenes \cite{shotton2013scene} with depth map labels.} We report the localization results on 7scenes when depth is available to refine the SfM models. The results are presented in cm, degree, and \% accuracy as in Table \ref{results_7scenes_table}.}
\label{7scenes_with_depth}
\begin{tabular}{c|c|c|c} 
\hline
\hline
                                                                        &                 & points            & points+lines                        \\ 
\hline
\multirow{2}{*}{\begin{tabular}[c]{@{}c@{}}without\\depth\end{tabular}} & Stairs          & 8.5 / 2.4 / 27.8  & 7.6 / 2.0 / 33.3                    \\
                                                                        & Avg. all scenes & 3.4 / 1.07 / 75.4 & 3.2 / 0.99 / 77.0                   \\ 
\hline
\multirow{2}{*}{\begin{tabular}[c]{@{}c@{}}with\\depth\end{tabular}}    & Stairs          & 8.4 / 2.5 / 35.7  & 4.7 / 1.24 / 53.0                   \\
                                                                        & Avg. all scenes & 3.2 / 1.05 / 79.6 & \textcolor{red}{2.6 / 0.85 / 83.4}  \\
\hline
\hline
\end{tabular}

\end{table}

\begin{table*}
\centering
\caption{\textbf{Localization results on the Cambridge Landmarks dataset \cite{kendall2015posenet}.} We present the median translation and rotation errors in cm and degrees and
pose accuracy (\%) at 5 cm / 5 deg. threshold of different relocalization methods using points and lines. The results in \textcolor{red}{red} are the best and \textcolor{blue}{blue} represents the second best in the same category of learning-based mapping method. The best overall results are in \textbf{bold}.}
\label{cambridge_table_results}

\begin{tabular}{c|c|c|c|c|c|c} 
\hline
\hline
                                    &                                                & Great Court                                             & King's College                      & Old Hospital                         & Shop Facade                          & St. Mary's Church                                      \\ 
\hline
\multirow{3}{*}{\rotcell{FM-\\based }}       & Hloc $^\text{point}$ \cite{sarlin2019coarse, sarlin2020superglue}                          & \textbf{9.5 / 0.05 / 20.4}                             & 6.4 / \textbf{0.10 / }37.0          & 12.5 / 0.23 / 22.5                   & 2.9 / 0.14 / 78.6                    & \textbf{3.7 /} 0.13 / 61.7                             \\
                                    & PtLine $^\text{point \& line}$ \cite{gao2022pose}                & 11.2 / 0.07 / 17.8                                     & 6.5 / \textbf{0.10 / }37.0          & 12.7 / 0.24 / 20.9                   & \textbf{2.7 / 0.12 /} 79.6           & 4.1 / 0.13 / 62.3                                      \\
                                    & Limap $^\text{point \& line}$ \cite{liu20233d}                 & 9.6 / 0.05 / 20.3                             & \textbf{6.2 / 0.10 / 39.4}          & \textbf{11.3 / 0.22 / 25.4}          & \textbf{2.7 /} 0.13 / \textbf{81.6}  & \textbf{3.7 / 0.12 / 63.8}                             \\ 
\hline
\multirow{6}{*}{\rotcell{Learning-\\based}} & SANet $^\text{dense}$ \cite{yang2019sanet}                          & 328 / 2.0 / -                                          & 32 / 0.5 / -                        & 32 / 0.5 / -                         & 10 / 0.5 / -                         & 16 / 0.6 / -                                           \\
                                    & DSAC* $^\text{dense}$ \cite{brachmann2021visual}                          & 49 / 0.3 / -                                           & 15 / 0.3 / -                        & 21 / 0.4 / -~                        & 5 / 0.3 / -                          & \textcolor{blue}{13 / 0.4 /} -                         \\
                                    & ACE $^\text{dense}$ \cite{brachmann2023accelerated}                            & 43 / 0.2 / -                         & 28 / 0.4 / -                        & 31 / 0.6 / -                         & 5 / 0.3 / -                           & 18 / 0.6 / -                                           \\
                                    & D2S $^\text{point}$ \cite{bui2023d2s}                            & 38 / 0.18 / -      & 15 / 0.24 / -                       & 21 / 0.40 / -                        & 6 / 0.32 / -                         & 16 / 0.50 / -                                          \\
                                    & PL2Map $^\text{point}$ (\textbf{ours})         & \textcolor{red}{33.0 / 0.16 /} \textcolor{blue}{2.51}                   & \textcolor{blue}{7.3 / 0.13 / 29.4} & \textcolor{blue}{16.8 / 0.30 / 8.34} & \textcolor{blue}{4.17 / 0.23 / 62.1} & \textcolor{red}{12.2 / 0.39 / 9.06}                    \\
                                    & PL2Map $^\text{point \& line}$ (\textbf{ours}) & \textcolor{blue}{33.8 /} \textcolor{red}{ 0.16 / 2.64} & \textcolor{red}{7.1 / 0.13 / 33.5}  & \textcolor{red}{15.4 / 0.28 / 9.34}  & \textcolor{red}{3.75 / 0.21 / 68.9}  & \textcolor{blue}{13.3 / 0.43 / }\textcolor{red}{9.06}  \\
\hline
\hline
\end{tabular}
\end{table*}

\subsection{Datasets and 3D Ground Truth Models}
We conducted our experiments on two standard camera re-localization datasets, both indoor and outdoor:

\textit{7Scenes} \cite{shotton2013scene}: An RGB-D dataset consists of seven different small environments, scaled up to 18$m^{3}$. For each scene, the authors provided several thousand images for training and testing. This dataset features challenges such as repeating structures of the Stair scene or motion blurs that occur throughout the dataset. Images were recorded using KinectFusion \cite{izadi2011kinectfusion} which also provides ground-truth poses and depth channels. 

\textit{Cambridge Landmarks} \cite{kendall2015posenet}: An RGB outdoor dataset scaling from 875$m^{2}$ to 8000$m^{2}$. The dataset consists of five scenes, including several hundred frames for the training and test sets, split by the authors. Ground-truth camera poses were obtained by reconstructing the SfM models. 

\textit{SfM ground truth models.} The main purpose of our method is to represent multiple 3D sparse models as a neural network for later localization with high-accuracy constraints and low storage demand. We leveraged two SfM tools, point-based SfM Hloc \cite{sarlin2019coarse, sarlin2020superglue} and line-based SfM Limap \cite{liu20233d} to obtain these models. Hloc and Limap are currently the most robust localization methods that utilize points and line maps. However, they are burdened by large memory requirements and complex feature-matching processes, posing challenges for real-time applications and lower-end consumer devices. This creates a large gap between real-time and low-grade consumer application. The proposed method addresses this burden. For all experiments, we used Superpoint \cite{detone2018superpoint} and 
DeepLSD \cite{pautrat2023deeplsd} to extract the 2D points and lines.

\subsection{Indoor Re-Mapping and Localization}
In this section, we report our localization results for learning indoor sparse point and line maps representation. We compare our results with those obtained from state-of-the-art baseline methods, as detailed in Table \ref{results_7scenes_table}. In Fig. \ref{predicted_3d_map}, we show some example results of 3D points and lines mapped by PL2Map.

We compare our indoor localization results against three established baselines: Hloc \cite{sarlin2019coarse}, PtLine \cite{gao2022pose}, and Limap \cite{liu20233d}. These were chosen because of their reliance on similar mapping components, specifically, points and lines. Regarding the use of only point maps, our approach demonstrated superior performance over Hloc in six out of seven evaluated scenes. For instance, within the Pumpkin scene, our method achieved localization errors of 3.5 cm in translation, 0.96$^{\circ}$ in rotation, and attained an accuracy of 63.1 \% (using a threshold of 5 cm / 5 deg). In contrast, Hloc exhibited less favorable results, with localization errors of 5.0 cm in translation, 1.32$^{\circ}$ of localization errors, and 50.4\% accuracy. These findings underscore the efficacy of our surrogate mapping model, which yields a performance improvement of 12.7\% over that of Hloc's database-based mapping approach. 

\textit{Line-assisted localization.} In Table \ref{results_7scenes_table}, we present our localization results using both the predicted points and line map. As can be seen, we achieved the best results when localizing with lines-assisted. These results confirm the efficiency of lines combined with points for visual localization, where an improvement can be observed for all seven scenes. Interestingly, our method learned using points and lines SfM models produced by Hloc and Limap, but the localization results obtained by our re-mapping method can even reach a large improvement margin. 

Additionally, we provide localization results with available depth to refine the SfM models in both point and line maps. The results are listed in Table \ref{7scenes_with_depth}. The original PL2Map still struggles with the repeated structure of the Stairs scene, but with SfM models refined using depth data, PL2Map shows a significant improvement in Stairs scene accuracy from 33.3\% to 53.0\%. The overall improvement was also observed for all scenes, evidenced in average median errors. Note that depth is used solely for training and inference relies solely on RGB images. 
\subsection{Outdoor Re-Mapping and Localization}
In this section, we present the localization results of the proposed method using the Cambridge outdoor dataset \cite{kendall2015posenet}. Unlike small-scale indoor environments, learning and SCR-based methods \cite{brachmann2021visual, bui2023d2s, brachmann2023accelerated} still struggle to close the gap with FM-based methods in large-scale outdoor scenarios. Thus, we compared our method with four additional SCR-based baselines: SANet \cite{yang2019sanet}, DSAC$^{*}$ \cite{brachmann2021visual}, ACE \cite{brachmann2023accelerated}, and D2S \cite{bui2023d2s}.  We present the results in Table \ref{cambridge_table_results}. Among the learning-based approaches, our localization results exhibited the lowest errors when utilizing only the predicted point maps. For instance, in the King’s College scene, our method outperforms \cite{bui2023d2s}, which also employs keypoint descriptors, by achieving a 51\% reduction in translation error. This substantial improvement narrows the accuracy gap with FM-based methods.

\textit{Line-assisted localization.} Table \ref{cambridge_table_results} also includes the results of our method when localization is achieved through the integration of both points and line maps. Across all five scenes, PL2Map demonstrated an enhancement in localization accuracy compared with scenarios where only the predicted points map was utilized. 

\begin{figure}
    \centering
    \includegraphics[width=0.45\textwidth]{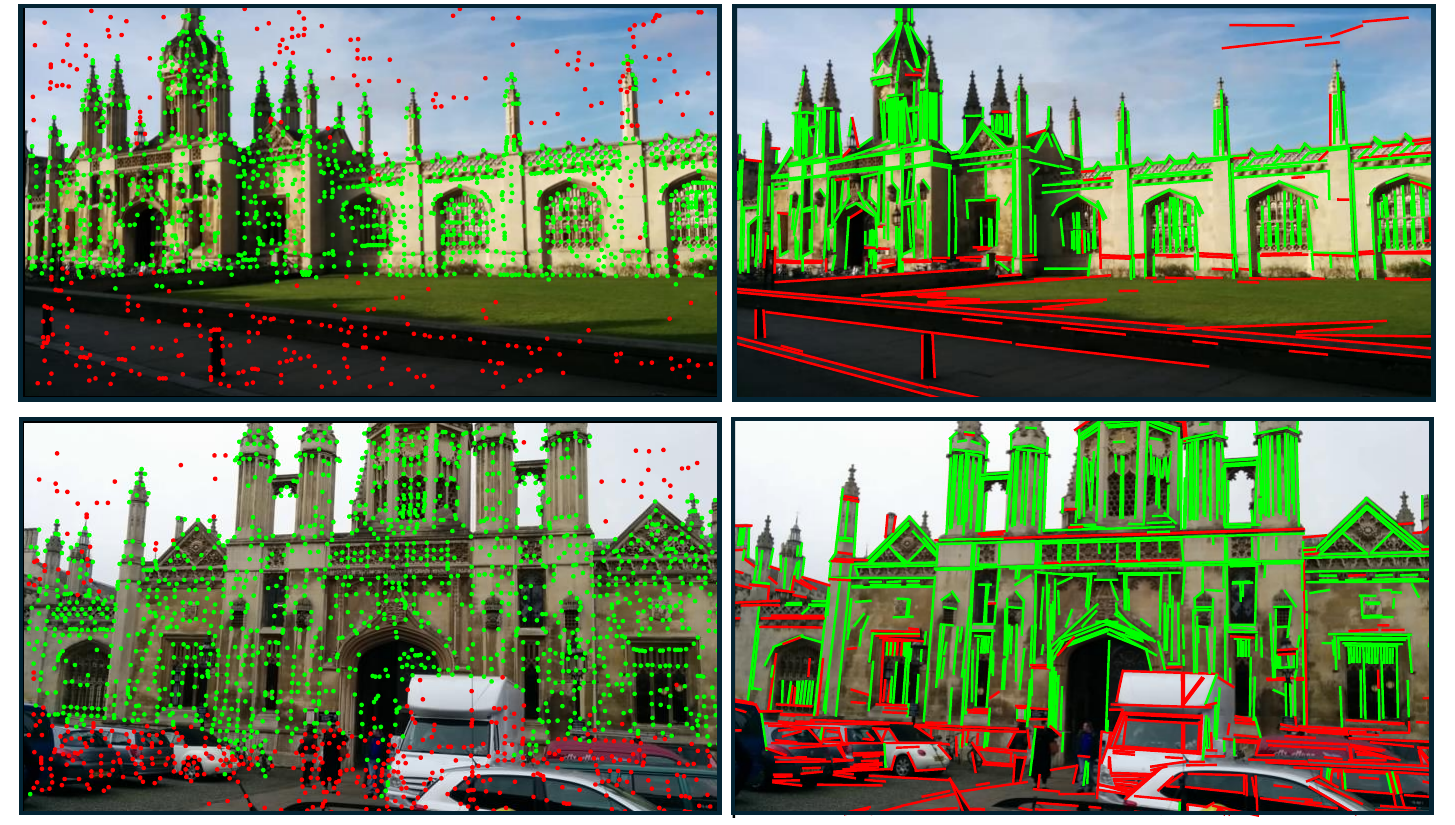}
    \caption{\textbf{Reliability prediction results on both 2D points and lines.} We show some examples of reliability prediction on the King's College scene \cite{kendall2015posenet} with thresholds of $r^{p}>0.6$ and $r^{l}>0.05$. }
    \label{2D_outliers}
\end{figure}

%%%%%%%%%%%%%%%%%%%%%%%%%%%%%%%%%%%%%%%%%%%%%%%%%%%%%%%%%%%%%%%%%%%%%%%%%%%%%%%%%

%%%%%%%%%%%%%%%%%%%%%%%%%%%%%%%%%%%%%

\subsection{Systems Efficiency}
Similar to \cite{bui2023d2s}, our method demonstrated the ability to disregard outlier features associated with dynamic elements or those unsuitable for localization through a single feedforward pass. This result is shown in Fig, \ref{2D_outliers}, where the PnP RANSAC step can benefit from focusing on a high-quality set of correspondences \cite{bui2023d2s}.  

Table \ref{re_quirement_localization} compares our approach with three primary baselines in terms of localization requirements, highlighting the efficiency of the proposed method by eliminating the need for a matching step and storing 3D maps as descriptors. Consequently, our approach requires significantly less memory, requiring approximately 25 MB for the network weight, in stark contrast to the several GBs required by Hloc \cite{sarlin2019coarse} and Limap \cite{liu20233d}.
\begin{table}
\centering
\caption{Comparison in requirements of localization pipeline using points and lines.}
\begin{tabular}{c|cccc} 
\hline\hline
\multirow{3}{*}{Requirements} & \multicolumn{4}{c}{Localization Method}  \\ 
\cline{2-5}
                              & Hloc & PtLine & Limap & Pl2Map           \\
                              & \cite{sarlin2019coarse}    & \cite{gao2022pose}      & \cite{liu20233d}     &  (ours)               \\ 
\hline
Database                      & yes  & yes    & yes   & no               \\
Image-retrieval               & yes  & yes    & yes   & no               \\
Points Matcher                & yes  & yes    & yes   & no               \\
Lines Matcher                 & -    & yes    & yes   & no               \\
\hline
\hline
\end{tabular}
\label{re_quirement_localization}
\end{table}

\section{CONCLUSIONS}
We propose the innovative PL2Map pipeline, designed to encapsulate sparse 3D points and lines within a unified model. After training with a designated scene, our pipeline efficiently generates 2D-3D correspondences for point and line features. In familiar settings, PL2Map not only serves as a cost-effective alternative to the conventional approach of storing and matching expensive descriptors but also shows robust re-mapping capabilities, which result in state-of-the-art camera relocalization. 

Future efforts could expand this work to a larger scale and include scene-agnostic pre-training of the attentional module across diverse conditions. Such advancements aim to achieve a quicker and more robust re-mapping methodology.


%%%%%%%%%%%%%%%%%%%%%%%%%%%%%%%%%%%%%%%%%%%%%%%%%%%%%%%%%%%%%%%%%%%%%%%%%%%%%%%%

% \clearpage
\bibliographystyle{unsrt}
\bibliography{reference.bib}

\end{document}